\documentclass[conference]{IEEEtran}
\IEEEoverridecommandlockouts
\usepackage{cite}
\usepackage{amsmath,amssymb,amsfonts}
\usepackage{algorithmic}
\usepackage{graphicx}
\usepackage{textcomp}
\usepackage{xcolor}
\usepackage{subcaption}
\usepackage{amsmath}
\usepackage{orcidlink}
\usepackage{hyperref}

\def\BibTeX{{\rm B\kern-.05em{\sc i\kern-.025em b}\kern-.08em
    T\kern-.1667em\lower.7ex\hbox{E}\kern-.125emX}}
\begin{document}

\title{AUV Acceleration Prediction Using DVL and Deep Learning}

\author{\IEEEauthorblockN{Yair Stolero and Itzik Klein \orcidlink{0000-0001-7846-0654}}
\IEEEauthorblockA{\textit{The Hatter Department of Marine Technologies} \\
\textit{Charney School of Marine Sciences, University of Haifa, Israel}\\}
}

\maketitle

\begin{abstract}
Autonomous underwater vehicles (AUVs) are essential for various applications, including oceanographic surveys, underwater mapping, and infrastructure inspections. Accurate and robust navigation are critical to completing these tasks. To this end, a Doppler velocity log (DVL) and inertial sensors are fused together. Recently, a model-based approach demonstrated the ability to extract the vehicle acceleration vector from DVL velocity measurements. Motivated by this advancement, in this paper we present an end-to-end deep learning approach to estimate the AUV acceleration vector based on past DVL velocity measurements. Based on recorded data from sea experiments, we demonstrate that the proposed method improves acceleration vector estimation by more than 65\% compared to the model-based approach by using data-driven techniques. As a result of our data-driven approach, we can enhance navigation accuracy and reliability in AUV applications, contributing to more efficient and effective underwater missions through improved accuracy and reliability.
\end{abstract}

\begin{IEEEkeywords}
Autonomous underwater vehicle, Doppler velocity log, inertial navigation system , data-driven, deep learning
\end{IEEEkeywords}

\section{Introduction}
Autonomous underwater vehicles (AUVs) are deployed in a wide range of underwater applications, including security operations \cite{hagen2003hugin}, underwater construction \cite{huang2017efficient, kondo2004navigation}, and aquaculture monitoring \cite{bao2020integrated}. A fundamental requirement for AUV operation is navigation, which involves continuously determining position, velocity, and orientation \cite{9101092}. However, the unique underwater environment presents significant challenges for precise navigation, necessitating specialized solutions \cite{stutters2008navigation}.

Most navigation systems rely on an inertial measurement unit (IMU) comprising three orthogonal accelerometers and three orthogonal gyroscopes. An inertial navigation system (INS) processes these inertial readings to estimate the vehicle’s movement in any environment, including underwater. However, due to inherent errors in the measurements of the inertial sensor, the INS solution accumulates drift over time \cite{titterton1997strapdown}. To mitigate this issue, additional sensor inputs are required to maintain an accurate and stable navigation solution.

Since global navigation satellite system (GNSS) signals do not propagate underwater due to electromagnetic signal attenuation, alternative methods must be employed. Acoustic navigation, which relies on transponder beacons \cite{lee2007pseudo}, and geophysical navigation, which leverages physical seabed features \cite{rice2004geophysical}, are commonly used for localization \cite{paull2013auv}. However, these techniques require predefined infrastructures or environmental characteristics to function effectively. Other common sensors include magnetometers for heading estimation and pressure sensors for depth measurement \cite{geng2010accuracy}.

Among these aiding sensors, the Doppler velocity log (DVL) is widely used in underwater navigation due to its accuracy, reliability, and self-sufficiency. Mounted on the underside of the AUV, the DVL transmits acoustic pulses toward the seafloor, measures the reflected signals, and determines the vehicle’s velocity vector. This method allows the DVL to operate independently, even in unexplored underwater regions \cite{miller2010autonomous, sanchez2020autonomous}. DVL data is typically integrated with INS measurements using nonlinear filtering techniques, most commonly the extended Kalman filter (EKF) \cite{karimi2013comparison}. In this fusion process, DVL velocity readings help refine INS state estimates and compensate for inertial sensor errors.

In the past years, researchers have increasingly investigated the application of machine learning and deep learning (DL) techniques to underwater navigation. These data-driven approaches offer significant advantages by improving the adaptability and robustness of navigation systems. Cohen and Klein~\cite{cohen2024inertial} provided an extensive review of deep learning methods applied to inertial sensing and sensor fusion, emphasizing their potential impact on underwater navigation.
One of the deep learning models is "NavNet," designed to integrate data from an attitude and heading reference system and a DVL. This model has demonstrated enhanced navigation accuracy and fault tolerance \cite{zhang2020navnet}. Similarly, Mu et al. \cite{mu2019end} introduced an end-to-end navigation framework utilizing a hybrid recurrent neural network, which fuses AHRS and DVL data. Topini et al. \cite{topini2020lstm} developed a Long short-term memory (LSTM) based dead reckoning method capable of estimating surge and sway velocities using past velocity estimates and temporal sequences of generalized forces.
Building on these advancements, Lv et al. \cite{lv2021position} proposed a hybrid gated recurrent neural network for position correction. Unlike traditional navigation methods, this approach operates without a predefined motion model, thereby mitigating modeling errors. Meanwhile, Liu et al. \cite{liu2022sins} introduced an INS/DVL integration technique that employs a radial basis function neural network to compensate for current-induced navigation inaccuracies.
Beyond those applications, deep learning models have been employed to address challenging scenarios where DVL functionality is impaired, such as sensor failures, partial beam loss, and measurement anomalies. Several studies have explored data-driven methods to enhance navigation reliability under these conditions \cite{yona2021compensating, davari2021real, li2021underwater, saksvik2021deep, cohen2023set, lv2020underwater, yona2024missbeamnet}. Additionally, BeamsNet, an end-to-end DL framework designed to estimate an AUV’s velocity vector in the absence of DVL data, was introduced in \cite{cohen2022beamsnet}.

A significant limitation of DVL-based navigation arises in scenarios where some or all of its acoustic beams are unavailable. A complete three-dimensional velocity solution requires at least three beams to reflect off the seafloor. Under challenging conditions, such as extreme roll or pitch angles, navigating over steep underwater terrain, or encountering large marine life, beam loss can occur, leading to gaps in velocity measurements.
Several strategies have been proposed to address these challenges \cite{yona2024missbeamnet, liu2018ins}. Recently, Klein and Lipman \cite{klein2020continuous,klein2022estimating} introduced an algorithm capable of estimating velocity during complete DVL outages by utilizing past DVL measurements. As part of this approach, they also estimated the AUV’s acceleration vector, though it remained unutilized in their final implementation.
Recently,  an approach to enhance the fusion between the INS and DVL by utilizing DVL-based acceleration measurements was proposed~\cite{levy2023ins}. They showed that the acceleration measurements improve the accuracy and convergence time of the standard INS/DVL fusion by an average improvement of $18\%$ for the straight-line trajectory and $40\%$ for the eight-figure trajectory. 

In this paper, we present a simple yet effective end-to-end deep learning approach for estimating the acceleration vector of an AUV. As in the baseline model-based approach, our method requires only past DVL measurements.  To validate our approach, we leveraged previously recorded sea experiment data, collected in the Mediterranean Sea near Haifa, Israel. These experiments were conducted using Haifa
University’s AUV, Snapir. The dataset includes inertial and DVL
measurements. By utilizing real-world marine data, we were able to thoroughly assess the effectiveness of our proposed method in practical
underwater navigation scenarios. The results demonstrate that
our approach achieves an improvement of up to $67.2\%$ in
estimating the acceleration vector compared to the model-based
approach. The proposed approach requires only additional software, making it suitable for operational AUVs. Furthermore, our approach allows the use of lower-grade sensors due to its improved overall performance. As a result, simple and low-cost navigation systems for small and micro platforms could be developed.

The rest of the paper is structured as follows: Section \ref{sec:proposed_approach} provides a detailed explanation of our proposed approach, Section \ref{sec:results} presents the research findings, and Section \ref{sec:conclusions} discusses the conclusions.

\section{Least Squares Acceleration Estimator} \label{sec:proposed_approach}
The model-based approach aims to compute the acceleration vector by using the velocity measurements obtained from the DVL over a series of time intervals. The analytical expression for estimating the AUV’s acceleration vector, denoted as $\tilde{\boldsymbol{a}^b}$, is derived as follows~\cite{klein2022estimating}:
\begin{equation}
    \tilde{\boldsymbol{a}^b} = \mathbf{B}^{-1} 
\begin{bmatrix}
 \sum_{i=0}^{n-1} \tilde{\boldsymbol{v}_j^b}  \\
 \sum_{i=0}^{n-1} \tilde{\boldsymbol{v}_j^b} \Delta_i 
\end{bmatrix}
\end{equation}
where $\tilde{\boldsymbol{a}^b}$ represents the estimated acceleration vector in the AUV's body frame, $\tilde{\boldsymbol{v}i^b}$ corresponds to the velocity measurements from the DVL at the $i{th}$ time step, and $\Delta_i$ is the time difference between the $i_{th}$ measurement and the first velocity measurement, defined as:
\begin{equation} \Delta_i = t_i - t_0 \end{equation}
where $t_i$ is the timestamp of the $i_{th}$ velocity measurement, and $t_0$ is the timestamp of the first velocity measurement. The time difference $\Delta_i$ is crucial because it represents the elapsed time between measurements, which is necessary for computing the velocity change over time and, consequently, estimating acceleration.

The matrix $\mathbf{B}$ is a key component in the least squares method and is expressed as:

\begin{equation} \mathbf{B} = \begin{bmatrix} \sum_{i=0}^{n-1} \Delta_i & \sum_{i=0}^{n-1} \Delta_i^2 \end{bmatrix} \end{equation}

The matrix $\mathbf{B}$ is constructed from the sums of the time differences $\Delta_i$ and their squares over the range of measurements. These terms are used in the least squares solution to obtain the best-fit acceleration estimate by minimizing the error in the velocity measurements. Essentially, this model-based approach attempts to derive a linear relationship between the velocities and the time intervals, which allows for the calculation of the acceleration vector over time.
\section{Proposed Approach} \label{subsel:dl_architecture}
\begin{figure}[h!] 
\begin{center} 
\captionsetup{justification=centering} \includegraphics[width=\columnwidth]{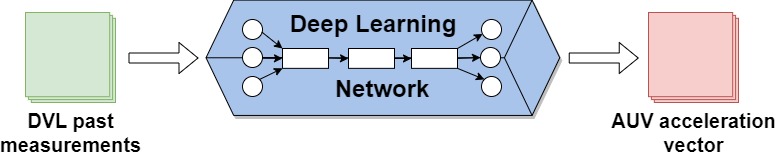} 
\caption{Our proposed AUV acceleration estimation approach.} \label{fig:proposed_approach} \end{center} 
\end{figure}

\begin{figure}[h!] \begin{center} \captionsetup{justification=centering} \includegraphics[width=\columnwidth]{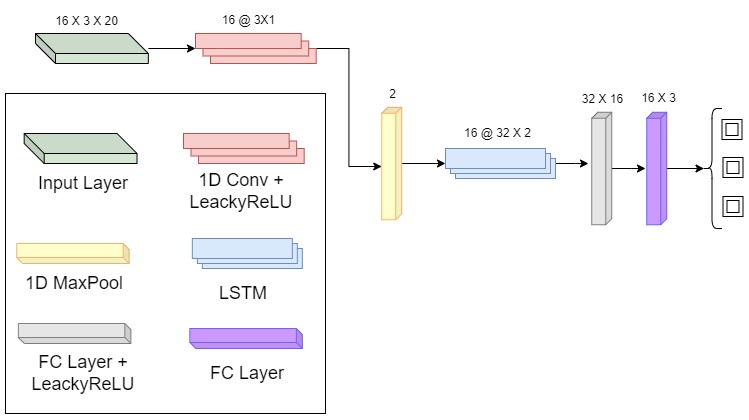} \caption{The  CNN-LSTM network architecture for predicting AUV acceleration.} \label{fig:network} \end{center} \end{figure}

This study focuses on estimating the AUV's acceleration vector by incorporating previous DVL measurements. To accomplish this, a deep learning architecture has been designed, combining the advantages of a convolutional neural network (CNN)  and a long short-term memory (LSTM) layers. A block diagram of the proposed approach is shown in Figure \ref{fig:proposed_approach}
CNN layers are utilized for their ability to efficiently extract local temporal features from time-series data, while the LSTM captures long-term dependencies and sequential patterns, enhancing the model's overall performance.
The layout of the proposed network architecture is illustrated in Figure \ref{fig:network}.
The model begins with an input layer that processes DVL velocity measurements formatted as a time series sequence. The input data consists of multiple past velocity vectors, enabling the network to learn temporal correlations in the motion dynamics.
\begin{equation}
    \mathbf{X} =\{ \mathbf{v_x},\mathbf{ v_y}, \mathbf{v_z} \} \in \mathbb{R}^{3 \times N}
\end{equation}
where $\mathbf{v_x, v_y, v_z}$ are the AUV velocity measurements along the x, y, and z axes, respectively, and N is the number of DVL measurements.
The first stage of the network is a 1D convolutional layer with a kernel size of three, followed by LeackyReLU activation function, specifically designed to identify localized temporal patterns in the velocity data, such as abrupt changes or smooth transitions. The 1D convolutional equation is:
\begin{equation} 
    y_f(t) =  \sum_{c=1}^{C} \sum_{i=1}^{k} w_f^{(c,i)} x_c (t+i-1)) + b_f
\end{equation}
where $y_f(t)$ is the output of the convolutional layer, $k$ represents the kernel size, $w_f$ are the learnable weights for filter $f$, channel $c$ and offset $i$, $x_c (t+i-1)$ is the input for channel $c$ at time stamp $t$, and $b_f$ is the bias.
The activation function is the LeakyReLU, defined as follows:
\begin{equation}
  \text{LeakyReLU}(y_f(t)) = \begin{cases} 
        y_f(t), & \text{if } Z_i \geq 0 \\
        \alpha \cdot y_f(t), & \text{otherwise}
        \end{cases}
\end{equation}
where \( \alpha \) (slope) is set to 0.01.

This layer is followed by a 1D MaxPooling layer with a pool size of three, which reduces the dimensionality of the feature map and highlights the most salient features, ensuring computational efficiency:
\begin{equation}
    y_f^P(t) = \max_{m=1}^{p} y_f (t+i)
\end{equation}
where $y_f^P(t)$ is the output of the ma pooling layer at time step t, $y_f (t+i)$ is the value of the feature map $f$ at time $t+i$ before pooling, and $p$ is the pooling size.

The feature representations from the CNN layer are then passed to the sequential modeling component, consisting of two stacked LSTM layers. Each layer contains 32 hidden units, enabling the network to retain and process information across longer time horizons:
\begin{align}
    f_t &= \sigma(W_f \cdot [h_{t-1}, Y_t] + b_f) \\
    i_t &= \sigma(W_i \cdot [h_{t-1}, Y_t] + b_i) \\
    \tilde{C}_t &= \tanh(W_C \cdot [h_{t-1}, Y_t] + b_C) \\
    C_t &= f_t \cdot C_{t-1} + i_t \cdot \tilde{C}_t \\
    o_t &= \sigma(W_o \cdot [h_{t-1}, Y_t] + b_o) \\
    h_t &= o_t \cdot \tanh(C_t)
\end{align}
where $f_t$ is the forget gate, $i_t$ is the input gate, $\tilde{C}_t$ is the candidate cell state, $C_t$ is the cell state at time $t$, $o_t$ is the output gate, $h_t$ is the hidden state at time $t$, $Y_t$ is the input at time $t$, $h_{t-1}$ is the previous hidden state, $C_{t-1}$ is the previous cell state, $W_f$, $W_i$, $W_C$, and $W_o$ are the weight matrices for the forget, input, candidate, and output gates respectively,  $b_f$, $b_i$, $b_C$, and $b_o$ are the bias terms for the gates, $\sigma$ is the sigmoid activation function, $\tanh$ is the hyperbolic tangent activation function.
This sequential processing is crucial for capturing complex temporal relationships inherent in AUV motion dynamics. 

The output from the LSTM layers is subsequently fed into two fully connected layers, which use LeakyReLU activation functions to produce the final predictions for the acceleration vector along the three axes (x, y, z):
\begin{equation}
    \boldsymbol{L_{FC_1}} = \text{LeakyReLU}(\boldsymbol{W_{FC_1}}h_t  + \boldsymbol{b_{FC_1}})
\end{equation}
where $\boldsymbol{W_{FC_1}}$ and $\boldsymbol{b_{FC_1}}$ are the weights and biases of the first fully connected layer, respectively. We applied the activation function and repeated the process for the second fully connected layer, resulting in the final output $\hat{y}_i$:
\begin{equation}
    \hat{y}_i = \boldsymbol{W_{FC_2}}\boldsymbol{L_{FC_1}} + \boldsymbol{b_{FC_2}}
\end{equation}
where $\boldsymbol{W_{FC_2}}$ and $\boldsymbol{b_{FC_2}}$ are the weights and biases of the second fully connected layer, respectively.

For the training process, we used the mean squared error (MSE) loss function:
\begin{equation}
\text{MSE} = \frac{1}{n} \sum_{i=1}^{n} ||y_i - \hat{y}_i||^2
\end{equation}
where $n$ is the number of data points, $y_i$ is the groud truth (GT) acceleration vector, and $\hat{y}_i$ is the predicted acceleration vector. 
The training was conducted over 30 epochs, with batch size of 16 and learning rate of 0.001.
\section{Dataset} \label{subsec:dataset}
\begin{figure}[h!] 
\begin{center} 
\includegraphics[width=0.8\columnwidth]{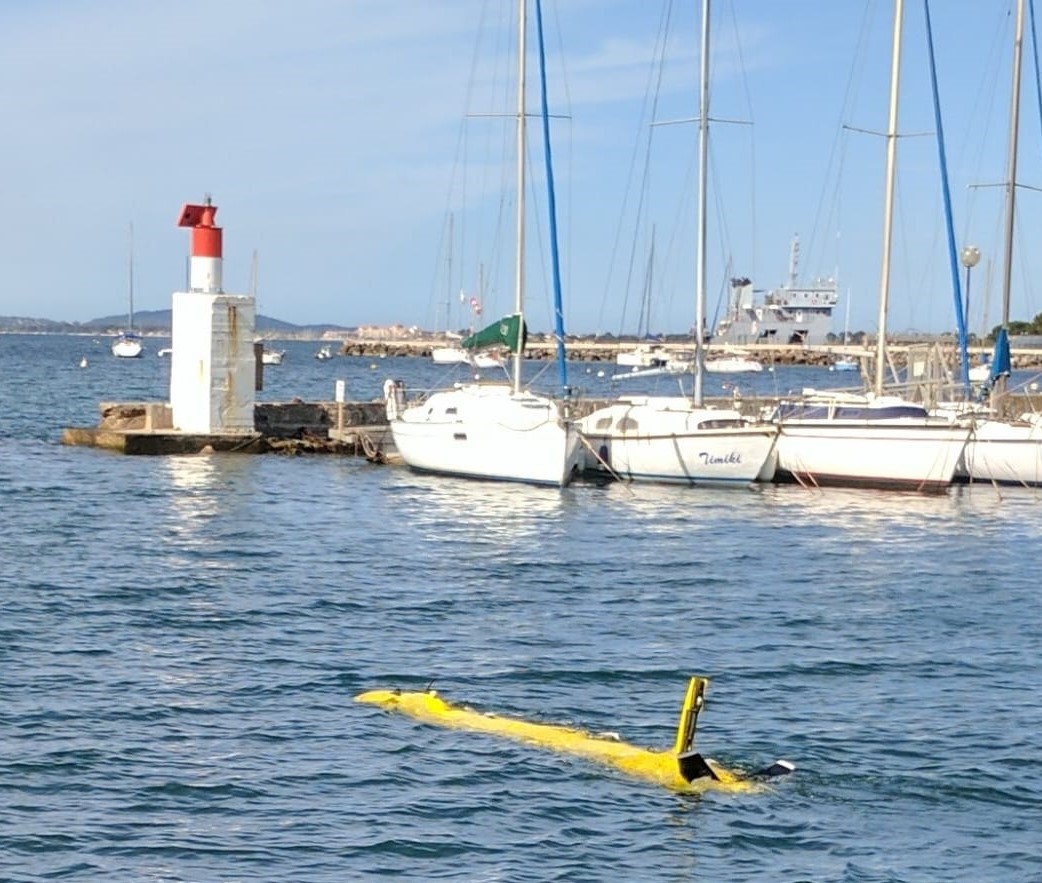}
\caption{The Snapir AUV during experimental trials.} \label{fig:experiment}
\end{center}
\end{figure}

\begin{figure}[h!] 
\begin{center} 
\includegraphics[width=0.8\columnwidth]{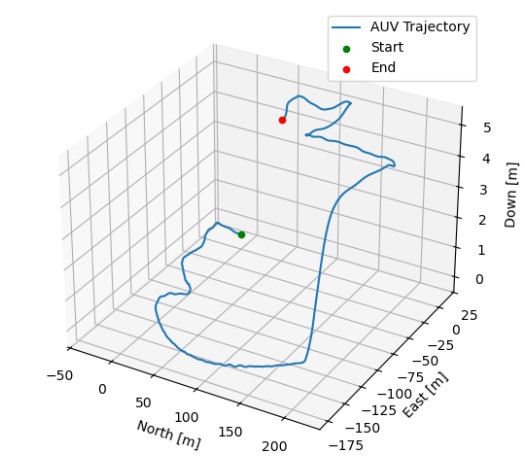}
\caption{A representative training trajectory made by the Snapir AUV.} \label{fig:tajectory1}
\end{center}
\end{figure}

The dataset used for this study was collected during a series of controlled experiments with the Snapir AUV in the Mediterranean Sea near Haifa, Israel~\cite{cohen2024kit}. The Snapir is a modified, ECA Robotics A18D mid-size AUV. It is equipped with a FOG based high performance inertial system operating at 100 Hz and a Teledybe RDI Work Horse navigattor DVL operating at 1 Hz. The DVL provides a velocity measurements with a standard deviation of 0.02 $m/s$ . Figure \ref{fig:experiment} presents the Snapir AUV during one of the experiments. \\
For training purposes, 66.7 minutes of data were extracted from the collected dataset, divided into 10 segments, each lasting 400 seconds. A A representative training trajectory is shown in Figure~\ref{fig:tajectory1}.
To enhance the robustness and generalization capability of the model, data augmentation techniques were employed. These included introducing artificial variations such as added scale, bias, and Gaussian white noise to added to the recorded data. We create ten additional noisy segments, that were added to the original ten segments. This process doubled our total training set to 133.4 minutes, divided into 20 segments.
%
%
The evaluation of the model was conducted using three additional 400 second segments that were excluded from the training dataset. These testing segments provide an unbiased assessment of the model's ability to predict acceleration accurately. The deliberate separation of training and testing data, along with the use of diverse operational scenarios, ensures that the proposed DL model's performance is both reliable and representative of real world applications.

Each dataset pair includes the ground truth (GT) acceleration vector and measured DVL velocity vector. The GT accelration required for training our DL model was derived directly using the GT position and velocity as obtained from the INS/DVL fusion process.
%
%
\section{Analysis and Results} \label{sec:results}
The  root mean squared error (RMSE) metric is used for our model evaluations. The RMSE is defined by:
\begin{equation}
    \text{RMSE} = \sqrt{\frac{1}{N} \sum_{i=1}^{N} ||\boldsymbol{y}_i - \hat{\boldsymbol{y}}_i||^2}
    \label{eq:rmse}
\end{equation}

where $N$ is the number of points in the dataset, $y_i$ represents the GT acceleration vector, and $\hat{y}_i$ is the estimated acceleration vector.
\begin{figure}[h!] \begin{center} \captionsetup{justification=centering} \includegraphics[width=\columnwidth]{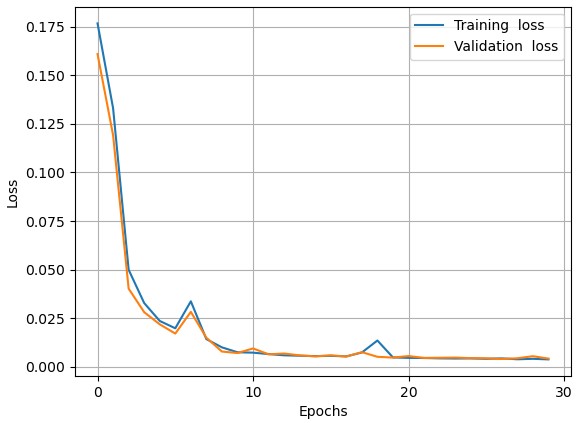} \caption{The training and validation loss graph versus the number of epochs.} \label{fig:val_loss} \end{center} \end{figure}
Figure \ref{fig:val_loss} shows how well our network is minimizing the training and validation loss over the training process. With the validation of the training part, we evaluate our proposed approach and compare it to the baseline approach~\cite{levy2023ins} using the test dataset. The comparison is presented in Table \ref{tab:results}. Our DL approach achieved a significant improvement in the acceleration prediction accuracy, reducing the RMSE by 67.2\%. This highlights the DL model's ability to capture the nonlinear and complex relationships inherent in AUV dynamics, which are often challenging for conventional methods. The results also underscore the DL model's adaptability across diverse operational conditions, including varying depths, speeds, and maneuvers. This adaptability is particularly evident in the consistent performance improvements observed during testing, demonstrating the model's robustness and reliability. 
\begin{table}[h!]
\centering
\captionsetup{justification=centering}
\caption{RMSE results of our approach and the model-based baseline for the test dataset.}
\begin{tabular}{|c|c|c|}
\hline
\begin{tabular}[c]{@{}c@{}}Our approach\\ RMSE $\frac{m}{s^2}$\end{tabular} & 
\begin{tabular}[c]{@{}c@{}}Model Based\\ RMSE $\frac{m}{s^2}$\end{tabular} & 
RMSE improvement {[}\%{]} \\ \hline
0.082 & 0.25 & 67.2 \\ \hline
\end{tabular}
\label{tab:results}
\end{table}
\section{Conclusions} \label{sec:conclusions}
In our research, we demonstrated that by utilizing past DVL velocity measurements, our proposed DL model outperforms the baseline model-based approach in estimating the AUV acceleration vector. The significant reduction in the acceleration RMSE underscores the model’s ability to effectively capture the intricate and nonlinear dynamics that govern AUV behavior. Unlike conventional methods, which rely on predefined motion models, the DL model adapts to the data, allowing it to better handle complex and variable underwater conditions.

One of the key strengths of the DL model is its ability to generalize across a range of different testing scenarios. Whether the AUV is operating at varying depths, speeds, or performing diverse maneuvers, the model maintains consistent and accurate performance. This flexibility makes the DL approach especially valuable in real-world applications where the operational conditions may change unpredictably.

With an accurate acceleration vector used as an additional measurement in the DVL/INS fusion process it is expected that the accuracy and convergence time of the standard INS/DVL would further improve. 
To conclude, our approach offers superior accuracy, adaptability, and real-time performance, all of which are crucial for advancing autonomous underwater exploration and operations.
%
\section{Acknowledgments}
Y.S. is grateful for the support of the University of Haifa Data Science Research Center.
\bibliographystyle{IEEEtran}
\bibliography{bib}

\begin{thebibliography}{10}
\providecommand{\url}[1]{#1}
\csname url@samestyle\endcsname
\providecommand{\newblock}{\relax}
\providecommand{\bibinfo}[2]{#2}
\providecommand{\BIBentrySTDinterwordspacing}{\spaceskip=0pt\relax}
\providecommand{\BIBentryALTinterwordstretchfactor}{4}
\providecommand{\BIBentryALTinterwordspacing}{\spaceskip=\fontdimen2\font plus
\BIBentryALTinterwordstretchfactor\fontdimen3\font minus \fontdimen4\font\relax}
\providecommand{\BIBforeignlanguage}[2]{{%
\expandafter\ifx\csname l@#1\endcsname\relax
\typeout{** WARNING: IEEEtran.bst: No hyphenation pattern has been}%
\typeout{** loaded for the language `#1'. Using the pattern for}%
\typeout{** the default language instead.}%
\else
\language=\csname l@#1\endcsname
\fi
#2}}
\providecommand{\BIBdecl}{\relax}
\BIBdecl

\bibitem{hagen2003hugin}
P.~E. Hagen, N.~Storkersen, K.~Vestgard, and P.~Kartvedt, ``The hugin 1000 autonomous underwater vehicle for military applications,'' in \emph{Oceans 2003. Celebrating the Past... Teaming Toward the Future (IEEE Cat. No. 03CH37492)}, vol.~2.\hskip 1em plus 0.5em minus 0.4em\relax IEEE, 2003, pp. 1141--1145.

\bibitem{huang2017efficient}
S.-W. Huang, E.~Chen, and J.~Guo, ``Efficient seafloor classification and submarine cable route design using an autonomous underwater vehicle,'' \emph{IEEE Journal of Oceanic Engineering}, vol.~43, no.~1, pp. 7--18, 2017.

\bibitem{kondo2004navigation}
H.~Kondo and T.~Ura, ``Navigation of an {AUV} for investigation of underwater structures,'' \emph{Control Engineering Practice}, vol.~12, no.~12, pp. 1551--1559, 2004.

\bibitem{bao2020integrated}
J.~Bao, D.~Li, X.~Qiao, and T.~Rauschenbach, ``Integrated navigation for autonomous underwater vehicles in aquaculture: A review,'' \emph{Information Processing in Agriculture}, vol.~7, no.~1, pp. 139--151, 2020.

\bibitem{9101092}
P.~Groves, \emph{Principles of GNSS, Inertial, and Multisensor Integrated Navigation Systems, Second Edition}.\hskip 1em plus 0.5em minus 0.4em\relax Artech, 2013.

\bibitem{stutters2008navigation}
L.~Stutters, H.~Liu, C.~Tiltman, and D.~J. Brown, ``{N}avigation technologies for autonomous underwater vehicles,'' \emph{IEEE Transactions on Systems, Man, and Cybernetics, Part C (Applications and Reviews)}, vol.~38, no.~4, pp. 581--589, 2008.

\bibitem{titterton1997strapdown}
D.~Titterton and J.~Weston, ``Strapdown inertial navigation,'' \emph{London: Peter Peregrinus Ltd}, 1997.

\bibitem{lee2007pseudo}
P.-M. Lee and B.-H. Jun, ``Pseudo long base line navigation algorithm for underwater vehicles with inertial sensors and two acoustic range measurements,'' \emph{Ocean Engineering}, vol.~34, no. 3-4, pp. 416--425, 2007.

\bibitem{rice2004geophysical}
H.~Rice, S.~Kelmenson, and L.~Mendelsohn, ``Geophysical navigation technologies and applications,'' in \emph{PLANS 2004. Position Location and Navigation Symposium (IEEE Cat. No. 04CH37556)}.\hskip 1em plus 0.5em minus 0.4em\relax IEEE, 2004, pp. 618--624.

\bibitem{paull2013auv}
L.~Paull, S.~Saeedi, M.~Seto, and H.~Li, ``{AUV} navigation and localization: A review,'' \emph{IEEE Journal of Oceanic Engineering}, vol.~39, no.~1, pp. 131--149, 2013.

\bibitem{geng2010accuracy}
Y.~Geng, R.~Martins, and J.~Sousa, ``Accuracy analysis of {DVL/IMU}/magnetometer integrated navigation system using different {IMU}s in {AUV},'' in \emph{IEEE ICCA 2010}.\hskip 1em plus 0.5em minus 0.4em\relax IEEE, 2010, pp. 516--521.

\bibitem{miller2010autonomous}
P.~A. Miller, J.~A. Farrell, Y.~Zhao, and V.~Djapic, ``Autonomous underwater vehicle navigation,'' \emph{IEEE Journal of Oceanic Engineering}, vol.~35, no.~3, pp. 663--678, 2010.

\bibitem{sanchez2020autonomous}
P.~J.~B. S{\'a}nchez, M.~Papaelias, and F.~P.~G. M{\'a}rquez, ``Autonomous underwater vehicles: Instrumentation and measurements,'' \emph{IEEE Instrumentation \& Measurement Magazine}, vol.~23, no.~2, pp. 105--114, 2020.

\bibitem{karimi2013comparison}
M.~Karimi, M.~Bozorg, and A.~Khayatian, ``A comparison of dvl/ins fusion by ukf and ekf to localize an autonomous underwater vehicle,'' in \emph{2013 First RSI/ISM International Conference on Robotics and Mechatronics (ICRoM)}.\hskip 1em plus 0.5em minus 0.4em\relax IEEE, 2013, pp. 62--67.

\bibitem{cohen2024inertial}
N.~Cohen and I.~Klein, ``Inertial navigation meets deep learning: A survey of current trends and future directions,'' \emph{Results in Engineering}, p. 103565, 2024.

\bibitem{zhang2020navnet}
X.~Zhang, B.~He, G.~Li, X.~Mu, Y.~Zhou, and T.~Mang, ``Navnet: Auv navigation through deep sequential learning,'' \emph{IEEE Access}, vol.~8, pp. 59\,845--59\,861, 2020.

\bibitem{mu2019end}
X.~Mu, B.~He, X.~Zhang, Y.~Song, Y.~Shen, and C.~Feng, ``End-to-end navigation for autonomous underwater vehicle with hybrid recurrent neural networks,'' \emph{Ocean Engineering}, vol. 194, p. 106602, 2019.

\bibitem{topini2020lstm}
E.~Topini, A.~Topini, M.~Franchi, A.~Bucci, N.~Secciani, A.~Ridolfi, and B.~Allotta, ``Lstm-based dead reckoning navigation for autonomous underwater vehicles,'' in \emph{Global Oceans 2020: Singapore--US Gulf Coast}.\hskip 1em plus 0.5em minus 0.4em\relax IEEE, 2020, pp. 1--7.

\bibitem{lv2021position}
P.-F. Lv, B.~He, and J.~Guo, ``Position correction model based on gated hybrid rnn for auv navigation,'' \emph{IEEE Transactions on Vehicular Technology}, vol.~70, no.~6, pp. 5648--5657, 2021.

\bibitem{liu2022sins}
P.~Liu, B.~Wang, G.~Li, D.~Hou, Z.~Zhu, and Z.~Wang, ``Sins/dvl integrated navigation method with current compensation using rbf neural network,'' \emph{IEEE Sensors Journal}, vol.~22, no.~14, pp. 14\,366--14\,377, 2022.

\bibitem{yona2021compensating}
M.~Yona and I.~Klein, ``Compensating for partial doppler velocity log outages by using deep-learning approaches,'' in \emph{2021 IEEE International Symposium on Robotic and Sensors Environments (ROSE)}.\hskip 1em plus 0.5em minus 0.4em\relax IEEE, 2021, pp. 1--5.

\bibitem{davari2021real}
N.~Davari and A.~P. Aguiar, ``Real-time outlier detection applied to a doppler velocity log sensor based on hybrid autoencoder and recurrent neural network,'' \emph{IEEE Journal of Oceanic Engineering}, vol.~46, no.~4, pp. 1288--1301, 2021.

\bibitem{li2021underwater}
D.~Li, J.~Xu, H.~He, and M.~Wu, ``An underwater integrated navigation algorithm to deal with dvl malfunctions based on deep learning,'' \emph{IEEE access}, vol.~9, pp. 82\,010--82\,020, 2021.

\bibitem{saksvik2021deep}
I.~B. Saksvik, A.~Alcocer, and V.~Hassani, ``A deep learning approach to dead-reckoning navigation for autonomous underwater vehicles with limited sensor payloads,'' in \emph{OCEANS 2021: San Diego--Porto}.\hskip 1em plus 0.5em minus 0.4em\relax IEEE, 2021, pp. 1--9.

\bibitem{cohen2023set}
N.~Cohen, Z.~Yampolsky, and I.~Klein, ``Set-transformer beamsnet for auv velocity forecasting in complete dvl outage scenarios,'' in \emph{2023 IEEE Underwater Technology (UT)}.\hskip 1em plus 0.5em minus 0.4em\relax IEEE, 2023, pp. 1--6.

\bibitem{lv2020underwater}
P.-F. Lv, B.~He, J.~Guo, Y.~Shen, T.-H. Yan, and Q.-X. Sha, ``Underwater navigation methodology based on intelligent velocity model for standard auv,'' \emph{Ocean Engineering}, vol. 202, p. 107073, 2020.

\bibitem{yona2024missbeamnet}
M.~Yona and I.~Klein, ``Missbeamnet: learning missing doppler velocity log beam measurements,'' \emph{Neural Computing and Applications}, vol.~36, no.~9, pp. 4947--4958, 2024.

\bibitem{cohen2022beamsnet}
N.~Cohen and I.~Klein, ``Beams{N}et: A data-driven approach enhancing {D}oppler velocity log measurements for autonomous underwater vehicle navigation,'' \emph{Engineering Applications of Artificial Intelligence}, vol. 114, p. 105216, 2022.

\bibitem{liu2018ins}
P.~Liu, B.~Wang, Z.~Deng, and M.~Fu, ``{INS/DVL/PS} tightly coupled underwater navigation method with limited {DVL} measurements,'' \emph{IEEE Sensors Journal}, vol.~18, no.~7, pp. 2994--3002, 2018.

\bibitem{klein2020continuous}
I.~Klein and Y.~Lipman, ``{C}ontinuous {INS/DVL} fusion in situations of {DVL} outages,'' in \emph{2020 IEEE/OES Autonomous Underwater Vehicles Symposium (AUV)}.\hskip 1em plus 0.5em minus 0.4em\relax IEEE, 2020, pp. 1--6.

\bibitem{klein2022estimating}
I.~Klein, Y.~Gutnik, and Y.~Lipman, ``Estimating {DVL} velocity in complete beam measurement outage scenarios,'' \emph{IEEE Sensors Journal}, vol.~22, no.~21, pp. 20\,730--20\,737, 2022.

\bibitem{levy2023ins}
O.~Levy and I.~Klein, ``{INS/DVL} fusion with {DVL} based acceleration measurements,'' \emph{arXiv preprint arXiv:2308.11762}, 2023.

\bibitem{cohen2024kit}
N.~Cohen and I.~Klein, ``Adaptive kalman-informed transformer,'' \emph{Engineering Applications of Artificial Intelligence}, vol. 146, p. 110221, 2025.

\end{thebibliography}
\end{document}